\documentclass[pre,twocolumn,amsmath,amssymb,aps,reprint,nofootinbib,superscriptaddress]{revtex4-2}

\usepackage{microtype}
\usepackage{graphicx}
\usepackage{booktabs}
\usepackage{subcaption}
\usepackage{float}
\usepackage{mathtools}
\usepackage{hyperref}
\usepackage[capitalize,noabbrev]{cleveref}

\begin{document}

\title{A Random-Matrix Criterion for Initializing\\
Gated Recurrent Neural Networks}

\author{Tommaso Fioratti}
\email{tommasofioratti@outlook.com}
\affiliation{Capital Fund Management, 75007 Paris, France}
\affiliation{Institute of Mathematics, EPFL, 1015 Lausanne, Switzerland}

\author{Riccardo Marcaccioli}
\affiliation{Capital Fund Management, 75007 Paris, France}

\author{Francesco Casola}
\affiliation{Capital Fund Management, 75007 Paris, France}

\begin{abstract}
Proper weight initialization prior to training has historically been one of the key factors that helped kick off the deep learning revolution. 
Initialization is even more crucial in ``reservoir computing'', where the weights of a readout layer are learned linearly while the reservoir weights are fixed and largely determine the richness, stability and memory of the resulting dynamics.
In the infinite-width limit it has been shown that meaningful initializations are those sitting at an effective critical point of the randomly initialized model. The phase transition is controlled by the weight variance $g^2$ and separates an ordered phase from a chaotic one where information progressively degrades. Here we derive a simple criterion to estimate the critical $g_c$ for a broad class of recurrent architectures and we show that it closely tracks the gain at which a gated-RNN reservoir achieves peak performance on a chaotic forecasting task. Finally, we argue that our criterion can serve as a design principle for future initialization schemes.
\end{abstract}

\maketitle

\section{Introduction}
Weight initialization has long been recognized as a key ingredient in the successful training of very deep models, as it helps preserve the scale of activations and gradients across layers and thereby mitigates the vanishing and exploding gradient problems \cite{Glorot2010, He2015}.

Early breakthroughs in large-scale deep learning \cite{Krizhevsky2012} demonstrated that substantial gains in classification performance could be achieved by combining new training and initialization schemes.

The initial theoretical justifications for the importance of initialization were largely heuristic, and weights were typically drawn randomly from small fixed-range distributions. The first systematic analyses focused on preserving the variance of forward signals and backward gradients, avoiding exponential distortion of information with depth \cite{Glorot2010, He2015}. Later work shifted the attention to the singular values of the full input--output Jacobian, linking optimal initialization with their concentration near unity \cite{Saxe2014, Pennington2017}.
At the same time, a vast literature on randomly initialized reservoirs showed that recurrent systems often achieve their best forecasting and information-processing performance when tuned close to the so-called ``edge of chaos'' (EOC) \cite{Bertschinger2004, Boedecker2012}.

From the perspective of statistical physics, the seminal work of Sompolinsky, Crisanti, and Sommers~\cite{Sompolinsky1988} was the first to quantitatively describe an EOC in a neural net using a mean-field continuous transition between a stationary and a chaotic phase in an effective many-body disordered glass-like system.
In their model, the variance $g^2$ of the normally initialized weights
$
U_{ij} \overset{\mathrm{i.i.d.}}{\sim} \mathcal{N}\!\left(0, \frac{g^2}{N}\right)
$
of the $N$-neuron synaptic matrix determines the stability of the zero fixed point. The critical $g_c$ could be obtained via a simple linearization of the equations of motion, studying the spectrum of the random matrix $U$.

More recently, in multilayer architectures, Schoenholz \emph{et al.}~\cite{Schoenholz2017} showed that randomly initialized deep networks are characterized by intrinsic depth scales that limit signal propagation through the architecture. Some of these depth scales, reminiscent of correlation lengths in physical systems \cite{Cardy1996}, diverge at criticality allowing the training of arbitrarily deep networks \cite{Pennington2017, Schoenholz2017, Saxe2014}.

Building on the observation that, in the Sompolinsky model, the loss of
stability of the trivial fixed point predicts the onset of chaos
\cite{Sompolinsky1988, Molgedey1992}, here we numerically demonstrate that
the same mechanism extends to a vast class of gated recurrent neural
networks (RNNs). In particular, we find that the correspondence between the
numerically estimated EOC and the fixed-point stability boundary persists
across a range of bias initializations. We leverage this connection to
define a practical criterion to compute $g_c$, studying the spectrum of
random matrices resulting from affine transformations of the synaptic
one~\cite{Fumarola2015}. While full random matrix theory and dynamical mean-field analyses of these architectures
have already been carried out in~\cite{Can2020, Krishnamurthy2022}, our
purpose is to bridge these theoretical developments and the everyday
choices faced by practitioners, by extracting a single closed-form
expression that can be evaluated directly from the bias initialization. 

As a concrete
illustration, we show that when a gated RNN is used as a dynamical
reservoir to predict a chaotic time series, optimal performance is obtained
by choosing the gain according to our criterion.

The paper is organized as follows. In Section~\ref{sec:model} we define the class of RNN models considered and show how common architectures are recovered as special cases. In Section~\ref{sec:eoc} we review numerical procedures to locate the EOC and we outline the assumptions underlying the critical $g_c$ predicted via a linearized model. In Section~\ref{sec:theory} we adapt the formalism introduced in \cite{Fumarola2015} to our linearized setting. In Section~\ref{sec:experiments} we validate the resulting criterion across architectures and initialization schemes through numerical experiments. In Section~\ref{sec:rc} we showcase the effectiveness of our framework in a reservoir-computing forecasting task. Finally, in Section~\ref{sec:discussion} we discuss limitations and open research directions.

\section{The Model}
\label{sec:model}
A recurrent neural network processes a sequence of inputs 
$\{x_t\}$ by maintaining a hidden state $\underline{h}_t \in \mathbb{R}^N$ 
that acts as an internal memory, encoding information about 
the past history of the sequence. At each time step, the 
network updates its hidden state and eventually produces an output 
through a parametric map as in the usual deep learning framework.
Different recurrent architectures, such as vanilla Recurrent Neural Networks (RNNs), Long Short-Term Memory networks (LSTMs) or Gated Recurrent Units (GRUs) 
differ in the specific form of the map, but they all share 
a common structure: the new hidden state is obtained by a nonlinear transformation
of the previous one and the external inputs, modulated by gating 
mechanisms of varying complexity. We exploit this observation 
to define a unified update rule that encompasses all these 
models as special cases.

Let $\underline{h}_{t} \in \mathbb{R}^N$ denote the hidden state at time $t$; the dynamics is defined as
\begin{equation}
\begin{aligned}
h_{t+1,i}
&= A_{t,i}\Bigl[(1-\alpha_{t,i})\,h_{t,i}
   + \alpha_{t,i}\,\phi\!\bigl(\hat{c}_{t+1,i}\bigr)\Bigr],
   \\
\hat{c}_{t+1,i}
&= g \sum_{j=1}^{N} U_{ij}\,o_{t,j}\,\Psi(h_{t,j})
   + \sum_{k=1}^{K} W_{ik}\,x_{t+1,k} + b_{c,i}.
\end{aligned}
\label{eq:general_gated_rnn}
\end{equation}
where:
\begin{itemize}
    \item $U$ is the synaptic weight matrix, $U_{ij} \overset{\mathrm{i.i.d.}}{\sim} \mathcal{N}\!\left(0, \frac{1}{N}\right)$; note that, compared with the Sompolinsky parameterization recalled in the Introduction, we factor the gain $g$ out of the weight distribution so that $g$ appears explicitly in~\eqref{eq:general_gated_rnn}.
    \item $W$ is the input weight matrix, $W_{ik} \overset{\mathrm{i.i.d.}}{\sim} \mathcal{N}\!\left(0, \frac{1}{K}\right)$.
    \item $x_{t} \in\mathbb{R}^K $ is the input at time $t$.
    \item $g$ is a global gain parameter controlling the strength of recurrent interactions.
    \item $\phi(\cdot)$ and $\Psi(\cdot)$ are element-wise nonlinear activation functions, e.g.\ $\tanh(x)$.
    \item $b_{c,i}$ is a bias in the candidate pre-activation, drawn as $b_{c,i} \overset{\mathrm{i.i.d.}}{\sim} \mathcal{N}(0, s_c^2)$.
    \item $o_{t,j}$ is an output (or reset) gate modulating the contribution of neuron $j$.
    \item $\alpha_{t,i} \in [0,1]$ is an update rate controlling the interpolation between memory retention and nonlinear update.
    \item $A_{t,i} > 0$ is an amplitude factor.
\end{itemize}
The update rule \eqref{eq:general_gated_rnn} can be interpreted as a 
nonlinear generalization of an exponential moving average (EMA)~\cite{Brown1956, kaufman1995}: 
a convex combination between the previous hidden state and a new 
candidate, modulated by an amplitude factor.
In the following, we always consider the autonomous case $x_t \equiv 0$ in order to study the properties of the network as a dynamical system.
The special cases recovered by the general dynamics~\eqref{eq:general_gated_rnn} are summarized in Table~\ref{tab:special_cases}.
\begin{table*}[t]
\centering
\begin{tabular}{lccccc}
\hline
\textbf{Model} 
& $A_{t,i}$ 
& $\alpha_{t,i}$ 
& $o_{t,i}$ 
& $\phi$
& $\Psi$ \\
\hline
RNN 
& $1$ 
& $1$ 
& $1$ 
& $\tanh$
& $\mathrm{Id}$ \\[0.3em]
LSTM 
& $f_{t,i} + i_{t,i}$ 
& $\displaystyle \frac{i_{t,i}}{f_{t,i} + i_{t,i}}$ 
& $o_{t,i}$ 
& $\tanh$
& $\tanh$ \\[0.6em]
GRU 
& $1$ 
& $z_{t,i}$ 
& $r_{t,i}$ 
& $\tanh$
& $\mathrm{Id}$ \\
\hline
\end{tabular}
\caption{Special cases of the general dynamics \eqref{eq:general_gated_rnn}.
Here $f_{t,i}$, $i_{t,i}$, and $o_{t,i}$ denote the forget, input, and output gates of the LSTM, while $z_{t,i}$ and $r_{t,i}$ denote the update and reset gates of the GRU.}
\label{tab:special_cases}
\end{table*}

In Table~\ref{tab:special_cases} the column $o_{t,i}$ is the recurrent-input modulator appearing in~\eqref{eq:general_gated_rnn}; its instantiation differs across architectures---unity in the vanilla RNN, the output gate $o_{t,i}$ in the LSTM, the reset gate $r_{t,i}$ in the GRU. The gates themselves ($f_{t,i}, i_{t,i}, o_{t,i}$ for LSTM and $z_{t,i}, r_{t,i}$ for GRU) are sigmoidal functions of the hidden state and the input,
\begin{equation}
\text{gate}_{t+1,i} = \sigma\!\Bigl(g\!\sum_{j} U^{\mathrm{g}}_{ij}\,h^{\mathrm{vis}}_{t,j} + \sum_{k} W^{\mathrm{g}}_{ik}\,x_{t+1,k} + b_{\mathrm{g}, i}\Bigr),
\label{eq:gate_form}
\end{equation}
with their own recurrent matrix $U^{\mathrm{g}}$, input matrix $W^{\mathrm{g}}$, and bias vector $b_{\mathrm{g}}$. The matrices $U^{\mathrm{g}}$ and $W^{\mathrm{g}}$ are sampled i.i.d.\ with the same distributions and gain $g$ as $U$ and $W$ in~\eqref{eq:general_gated_rnn}, while the bias entries are drawn $b_{\mathrm{g}, i} \overset{\mathrm{i.i.d.}}{\sim} \mathcal{N}(0, s_b^2)$. The visible state $\underline{h}^{\mathrm{vis}}_t$ presented to the gates is architecture-dependent:
\[
\underline{h}^{\mathrm{vis}}_t =
\begin{cases}
\underline{h}_t & \text{vanilla RNN and GRU},\\
o_t\,\tanh(\underline{h}_t) & \text{LSTM}.
\end{cases}
\]
In all three cases $\underline{h}^{\mathrm{vis}}_t$ vanishes at $\underline{h}_t = 0$, so at the autonomous fixed point $\underline{h} = 0$ every gate collapses to $\sigma(b_{\mathrm{g}, i})$---a property we repeatedly exploit in Section~\ref{sec:theory}.

\section{Finding the EOC}
\label{sec:eoc}

Our analysis starts by considering the autonomous free evolution 
with $x_t \equiv 0$ and the stability of the model dynamics as 
a function of the gain parameter~$g$ and the bias standard 
deviations~$s_b$ and~$s_c$, where~$s_b$ controls the gate biases 
and~$s_c$ the candidate bias in~\eqref{eq:general_gated_rnn}.

In many-body physics, a common strategy to determine phase-diagram 
instabilities is to look for an effective low-energy---i.e.\ 
long-time-scale---description of the system~\cite{Cardy1996}. 
Retaining the same philosophy, we consider the free evolution 
from the initial state $\underline{h}_0$ at times 
$t \gg 1$. Specifically, we study the quantity
\[
  q_t = \frac{1}{N}\sum_{i=1}^{N} h_{i,t}^2, 
  \qquad 
  q_{\infty} = \lim_{t \to \infty} q_t,
\]
which is reminiscent of the Edwards--Anderson order 
parameter~\cite{EdwardsAnderson1975} in glassy systems and has 
been studied by several authors in the context of random 
feedforward networks~\cite{Schoenholz2017, 
Poole2016}. Since $q_t$ is a mean-square quantity, 
$q_\infty = 0$ can only occur if $h_{i,t} \to 0$ for every 
neuron, making the origin a global attractor.

\begin{figure}[t]
  \centering

  \begin{subfigure}[t]{\linewidth}
      \centering
      \includegraphics[width=0.95\linewidth]{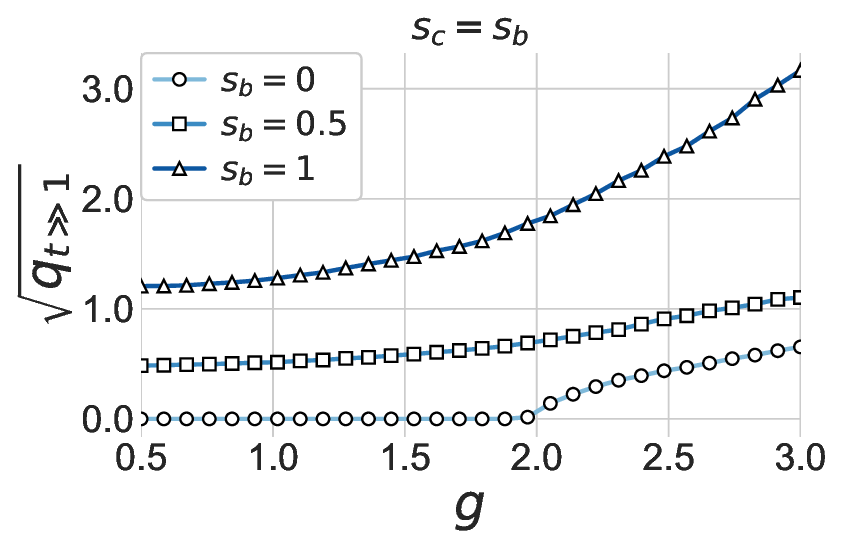}
  \end{subfigure}

  \vspace{0.8em}

  \begin{subfigure}[t]{\linewidth}
      \centering
      \includegraphics[width=0.95\linewidth]{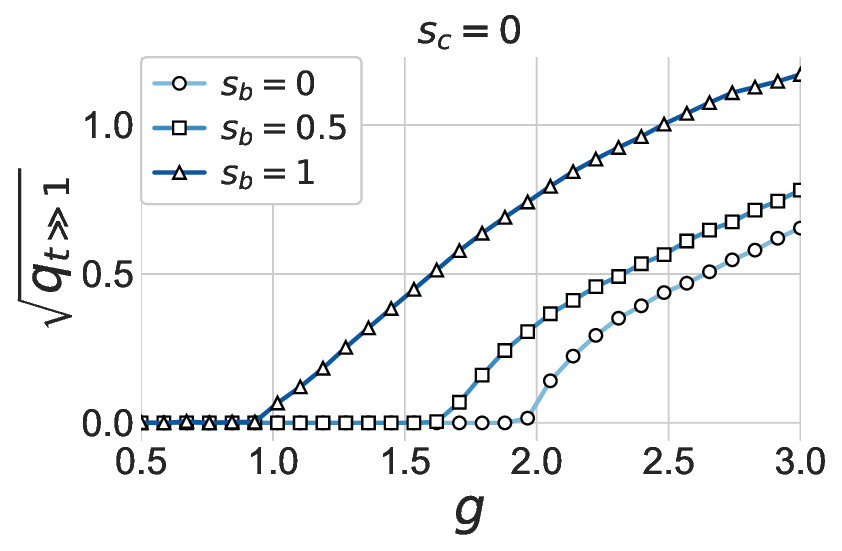}
  \end{subfigure}

  \caption{Stationary signal power in LSTM with Gaussian biases. 
  Comparison between including (Top) versus omitting 
  (Bottom) the bias in the candidate update. Hyperparameters: initial state
$\underline{h_0} = \mathbf{1}$,
steps $T = 4000$, neurons $N = 2000$, average over $R = 250$
replicas.}
  \label{fig:q-comparison}
\end{figure}
Figure~\ref{fig:q-comparison} reports estimates of $q_\infty$ as a function of the gain $g$, obtained by evolving the autonomous dynamics from a fixed initial condition for a long time $t \gg 1$ and averaging $q_{t \gg 1}$ over independent realizations of the quenched disorder.
The top and bottom panels differ 
only in whether $s_c$ is set to zero or not, and the two cases 
exhibit qualitatively different behavior.
For $s_c = 0$, $\underline{h} = 0$ is a fixed point of the dynamics in every realization of the quenched biases, and the evolution linearized around it reads $\underline{h}_{t+1} = J\,\underline{h}_t$, which is invariant under the discrete symmetry $\underline{h} \to -\underline{h}$. For $s_c > 0$ instead, the candidate bias displaces the fixed point away from the origin, acting as a random field in a system with quenched disorder~\cite{EdwardsAnderson1975, Sompolinsky1988}. As a consequence $q_\infty > 0$ even below $g_c$, and the linearization-based analysis of Section~\ref{sec:theory} does not directly apply. We include the $s_c = s_b$ case in Figure~\ref{fig:q-comparison} for completeness; in the rest of the paper we focus on $s_c = 0$, which preserves $\underline{h} = 0$ as an exact fixed point and is therefore amenable to the linearization-based analysis of Section~\ref{sec:theory}.

In what follows we refer to the sub-critical regime $g < g_c(s_b)$ 
with $s_c = 0$ as the \emph{ordered phase}, in which the origin 
acts as a global attractor of the dynamics for any $s_b \geq 0$.
The instabilities of the ordered phase can be studied by 
expanding the equations of motion around $h \approx 0$ in the 
$g \to g_c$ limit, obtaining an effective long-time evolution. 
We propose the use of random matrix theory to determine the 
instability of the ordered phase ($s_c = 0$) for the general 
system in~\eqref{eq:general_gated_rnn}, and leave the extension 
to the biased setting $s_c > 0$, where $q_\infty$ no longer 
vanishes, to future work.

To gain further insight into the ordered-to-chaotic transition 
at~$g_c$, we estimate the maximal Lyapunov exponent using the 
Benettin algorithm~\cite{Benettin1980} and combine it 
with a bisection search over~$g$. This allows us to identify the 
critical gain~$g_c$ at which the maximal Lyapunov exponent 
crosses zero.

\begin{figure}[t]
  \centering

  \begin{subfigure}[t]{\linewidth}
      \centering
      \includegraphics[width=0.9\linewidth]{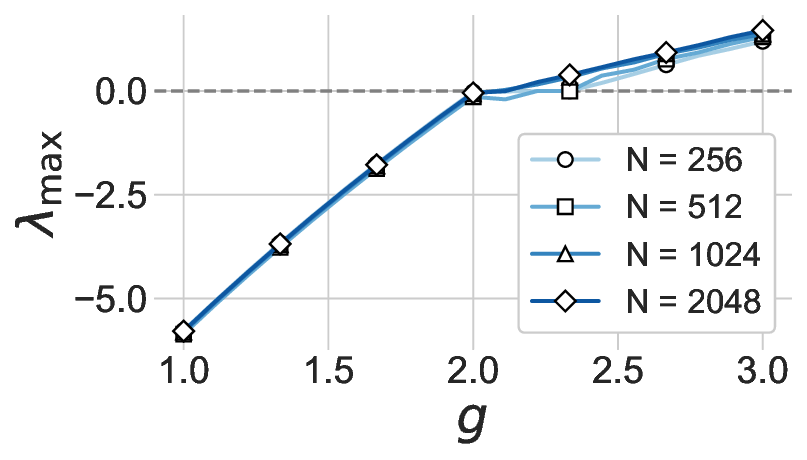}
  \end{subfigure}

  \vspace{0.8em}

  \begin{subfigure}[t]{\linewidth}
      \centering
      \includegraphics[width=0.9\linewidth]{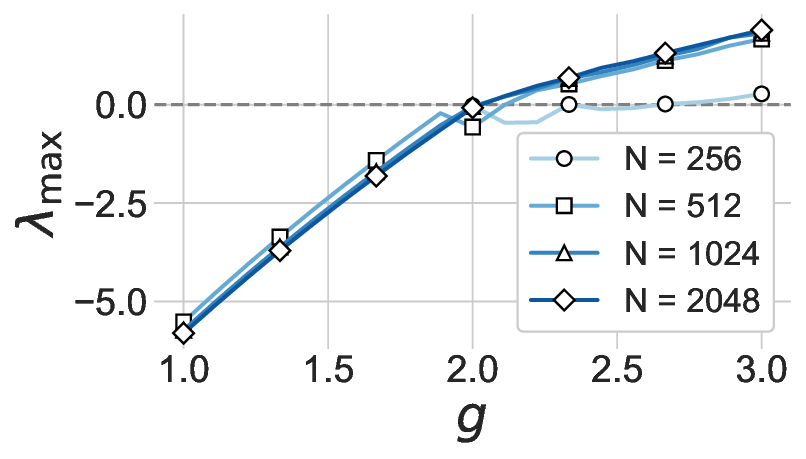}
  \end{subfigure}

  \caption{Maximum Lyapunov exponent $\lambda_{\max}$ as a function of the gain $g$, estimated via the Benettin algorithm~\cite{Benettin1980}, in the zero-bias case ($s_b = s_c = 0$) for LSTM (top) and GRU (bottom) at different hidden sizes $N$. For both architectures $\lambda_{\max}$ crosses zero at $g \simeq 2$, in agreement with the prediction $g_c = 2$ derived in Section~\ref{sec:theory}.}
  \label{fig:lambda-comparison}
\end{figure}

In Figure~\ref{fig:lambda-comparison} we can clearly see, e.g., 
that the estimated phase transition occurs at $g = 2$ for both 
LSTM and GRU when no bias is present in any of the gates 
($s_b = 0$).

\section{Theory}
\label{sec:theory}
Our strategy is to approach the phase transition from below, i.e.\ from the regime where the trivial fixed point is a global attractor (see Figure~\ref{fig:q-comparison}). In this setting, the dynamics can be studied through its linearization around $\underline{h}=0$, which yields an analytically tractable problem as long as trajectories remain in a neighborhood of the fixed point. The Jacobian evaluated at $\underline{h} = 0$ is a non-Hermitian random matrix; to characterize its spectral properties we rely on the theory originally developed in~\cite{Fumarola2015} to study partially random linear networks.

In a neighborhood of $\underline{h}=0$, the dynamics in~\eqref{eq:general_gated_rnn} can be linearized and written in matrix form as
\[
\underline{h}_{t+1} \approx D_{A^*}\big[ (I - D_{\alpha^*}) + D_{\alpha^*}\,g\,U\,D_{o^*} \big]\,\underline{h}_{t},
\]
where $D_{(\,\cdot\,)}$ denotes the diagonal matrix whose entries are given by the corresponding amplitude factor or gate evaluated at $\underline{h}=0$. In writing this expression we have used that $\phi'(0) = \Psi'(0) = 1$, which holds for all nonlinearities in Table~\ref{tab:special_cases}; the extension to generic activations amounts to inserting a factor $\phi'(0)$ in front of $D_{A^*}D_{\alpha^*}$ and a factor $\Psi'(0)$ in front of $D_{o^*}$ in the expressions below.

Thus, the stability of the fixed point is determined by the spectrum of the non-Hermitian random matrix
\begin{equation}
J = D_{A^*}(I - D_{\alpha^*}) + D_{A^*}D_{\alpha^*}\,g\,U\,D_{o^*}.
\label{eq:jacobian}
\end{equation}
Here $U$ is the synaptic weight matrix, whose empirical spectrum converges to the circular law of unit radius as $N\to\infty$~\cite{Tao2010}. The diagonal factors $D_{A^*}$, $D_{\alpha^*}$ and $D_{o^*}$ act as anisotropic scalings, yielding a deformed version of the circular-law spectrum for $J$.

In the case of LSTM and GRU architectures, the ``deformation'' terms $D_{A^*}(I - D_{\alpha^*})$, $D_{A^*}D_{\alpha^*}$, and $D_{o^*}$ are all diagonal matrices whose entries are of the form $\sigma(b_i)$ or $1 - \sigma(b_i) = \sigma(-b_i)$, where $
\sigma(x) = \frac{1}{1+ e^{-x}}$
is the sigmoid function and $b_i$ is the $i$-th bias of the corresponding gate (the explicit entries are given in Table~\ref{tab:MLR}). In particular, these matrices are diagonal with entries in $(0,1)$, hence invertible and positive definite.

Once the biases are sampled, $J$ splits into a deterministic diagonal part $D_{A^*}(I-D_{\alpha^*})$, with eigenvalues in $[0,1)$ for the architectures of Table~\ref{tab:special_cases}, plus a random non-Hermitian perturbation $g\,D_{A^*}D_{\alpha^*}\,U\,D_{o^*}$ driven by $U$. For $g=0$ the spectrum lies strictly inside the unit disk and $\underline{h}=0$ is linearly stable; as $g$ increases, the random term deforms the spectrum, eventually reaching the unit circle. In the $N\to\infty$ limit the spectrum of $J$ concentrates on a deterministic set characterized by the results of~\cite{Fumarola2015}; we define the critical gain $g_c$ as the smallest $g$ for which the boundary of this set touches the unit circle.

Adopting the notation of~\cite{Fumarola2015}, we set
\[
M = D_{A^*}(I - D_{\alpha^*}), \qquad
L = D_{A^*}D_{\alpha^*}, \qquad
R = D_{o^*}.
\]
The result of~\cite{Fumarola2015} characterizes, under appropriate regularity conditions, the limiting eigenvalue support of $J = M + g\,L\,U\,R$ for deterministic sequences of $M, L, R$ and a random matrix $U$ with iid entries. In our setting $M, L, R$ are themselves random through the gate biases. We apply the result of~\cite{Fumarola2015} along bias realizations, treating the iid bias structure as the mechanism that ensures convergence of the relevant empirical distributions; under this working interpretation, the limiting eigenvalue support is identified with the set of $z\in\mathbb{C}$ such that $K(0^+, z)\geq 1$, where
\begin{equation}
K(0^+,z) \;=\; \lim_{r\to 0^+}\lim_{N\to\infty}\frac{1}{N}\sum_{i=1}^{N}\frac{g^2}{\mathrm{sv}_i(z)^2 + r^2},
\label{eq:fumarola_condition}
\end{equation}
and $\mathrm{sv}_i(z)$ are the singular values of $L^{-1}(z-M)R^{-1}$. We do not attempt a full verification of the technical conditions of~\cite{Fumarola2015} in this random-bias setting; the agreement between the predicted $g_c$ and the numerical estimates reported in Section~\ref{sec:experiments} provides empirical support for this approach.

The regularizer $r$ in~\eqref{eq:fumarola_condition} is needed in~\cite{Fumarola2015} to handle deterministic sequences of $M, L, R$ for which the unregularized empirical sum may diverge in the large-$N$ limit. We argue in Appendix~\ref{app:reg} that, in the iid-bias setting and under the same working interpretation as above, this divergence does not occur for the cases considered in this work, and the regularizer can be omitted. The boundary of the spectral support is the locus where the inequality in~\eqref{eq:fumarola_condition} is saturated, so setting $r=0$ and imposing equality yields
\begin{equation}
    \frac{1}{N} \sum_{i=1}^{N} \frac{g^2}{\mathrm{sv}_i(z)^2} = 1.
\label{eq:fumarola_condition_equality}
\end{equation}

Since the matrices $L$, $R$ and $(zI-M)$ are diagonal, and $L, R$ are invertible and positive definite, the matrix $L^{-1}(zI-M)R^{-1}$ is itself diagonal with $i$-th entry $(z-M_{ii})/(L_{ii}R_{ii})$; its singular values are therefore
\begin{equation}
    \mathrm{sv}_i = \frac{|z - M_{ii}|}{L_{ii}\,R_{ii}}.
    \label{eq:sv_explicit}
\end{equation}
Plugging into~\eqref{eq:fumarola_condition_equality} and rearranging gives, for each $z$ on the spectral boundary,
\begin{equation}
    \frac{1}{g^2} = \frac{1}{N}\sum_{i=1}^{N} \frac{L_{ii}^2 R_{ii}^2}{|z - M_{ii}|^2}.
\end{equation}
Since increasing $g$ enlarges the spectral support, the critical gain $g_c$ corresponds to the maximum of the right-hand side over $|z|=1$:
\begin{equation}
    \frac{1}{g_c^2} = \max_{|z| = 1}\frac{1}{N}\sum_{i=1}^{N} \frac{L_{ii}^2 R_{ii}^2}{|z - M_{ii}|^2}.
\end{equation}
Since the $M_{ii}$ are real and lie in $[0,1)$, for every $z = x+iy$ with $|z|=1$ we have
\[
|z - M_{ii}|^2 = 1 - 2x\,M_{ii} + M_{ii}^2,
\]
which is minimized term-by-term at $z = 1$ (i.e.\ $x=1$, $y=0$). Since $L_{ii}, R_{ii} > 0$, the whole sum is therefore maximized at $z = 1$, giving
\begin{equation}
    g_c = \bigg(\frac{1}{N}\sum_{i=1}^{N} \frac{L_{ii}^2 R_{ii}^2}{(1 - M_{ii})^2}\bigg)^{-\frac{1}{2}},
    \label{eq:critical_g}
\end{equation}
a practical rule for computing the critical gain as a function of the bias. To apply it, the only ingredients needed are the diagonal entries $M_{ii}, L_{ii}, R_{ii}$. As shown in Section~\ref{sec:model}, at $\underline{h}=0$ every gate of~\eqref{eq:gate_form} collapses to $\sigma(b_{\mathrm{g}, i})$; substituting into Table~\ref{tab:special_cases} yields the entries reported in Table~\ref{tab:MLR} for LSTM and GRU.

\begin{table}[t]
\centering
\begin{tabular}{lccc}
\hline
& $M_{ii}$ & $L_{ii}$ & $R_{ii}$ \\
\hline
LSTM & $\sigma(b_{f,i})$ & $\sigma(b_{i,i})$ & $\sigma(b_{o,i})$ \\[3pt]
GRU  & $1-\sigma(b_{z,i})$ & $\sigma(b_{z,i})$ & $\sigma(b_{r,i})$ \\
\hline
\end{tabular}
\caption{Entries of the diagonal matrices $M = D_{A^*}(I - D_{\alpha^*})$, $L = D_{A^*}D_{\alpha^*}$ and $R = D_{o^*}$ for LSTM and GRU, obtained by evaluating the gates of Table~\ref{tab:special_cases} at the fixed point $\underline{h}=0$. Here $b_{\bullet,i}$ denotes the $i$-th entry of the bias vector of the corresponding gate. Plugging these into~\eqref{eq:critical_g} yields a closed-form expression for $g_c$ as a function of the bias initialization.}
\label{tab:MLR}
\end{table} 

\section{Application to gated RNNs}
\label{sec:experiments}
In this section we compare the numerically estimated phase transition with the criterion given by~\eqref{eq:critical_g} under three representative bias initializations.

\subsection{Zero bias}
\label{sec:zero_bias}

We start with the LSTM and GRU architectures in the zero-bias case.
Setting all biases to zero, the diagonal matrices $D_{A^*}(I-D_{\alpha^*})$, $D_{A^*}D_{\alpha^*}$, and $D_{o^*}$ have all entries equal to $\sigma(0) = 1/2$, so that
\[
M_{ii} = L_{ii} = R_{ii} = \tfrac{1}{2} \quad \forall\, i.
\]
The Jacobian at $\underline{h} = 0$ then reads
\[
J = \tfrac{1}{2}\,I + \tfrac{g}{4}\,U,
\]
and~\eqref{eq:critical_g} gives
\[
g_c
= \left(\frac{(1/2)^2\,(1/2)^2}{(1 - 1/2)^2}\right)^{-1/2}
= \left(\tfrac{1}{4}\right)^{-1/2}
= 2.
\]
Throughout the paper, when we write $g_c$ without further qualification we refer to this zero-bias value
\[
g_c \;\equiv\; g_c(s_b = 0) = 2.
\]
When the bias dependence is relevant, we write $g_c(s_b)$ explicitly.

As already shown in Figure~\ref{fig:lambda-comparison}, the fixed-point stability criterion is in very good agreement with the onset of chaos for both LSTMs and GRUs.

\subsection{Gaussian bias initialization}
\label{sec:gauss_bias}

We now consider a Gaussian bias initialization, where all biases are drawn from $\mathcal{N}(0, s_b^2)$.
Substituting the entries of Table~\ref{tab:MLR} into~\eqref{eq:critical_g} yields
\begin{equation}
g_c(s_b)
=
\left(
\frac{1}{N}\sum_{i=1}^{N}
\frac{\sigma(b_{L,i})^2\,\sigma(b_{R,i})^2}
{\bigl(1-\sigma(b_{M,i})\bigr)^2}
\right)^{-\frac{1}{2}}\!,
\label{eq:g_gauss}
\end{equation}
with $b_{\bullet,i} \sim \mathcal{N}(0, s_b^2)$ for $\bullet\in\{L,M,R\}$, the identification with each architecture's bias vectors being read off Table~\ref{tab:MLR}.

For the GRU, this expression simplifies considerably.
Since $L = I - M$ holds by construction (cf.\ Table~\ref{tab:MLR}), the ratio
\[
\frac{\sigma(b_{L,i})^2}{\bigl(1-\sigma(b_{M,i})\bigr)^2} = 1
\]
identically, and~\eqref{eq:g_gauss} collapses to
\begin{equation}
g_c(s_b) = \left(\frac{1}{N}\sum_{i = 1}^N R_{ii}^2\right)^{-1/2} \xrightarrow[N\to\infty]{} \langle \sigma(b)^2 \rangle^{-1/2}\!,
\label{eq:gc_gru_gauss}
\end{equation}
so that the GRU critical gain depends only on the reset-gate statistics.

\begin{figure}[t]
  \centering
 
  \begin{subfigure}[t]{\linewidth}
    \centering
    \includegraphics[width=0.85\linewidth]{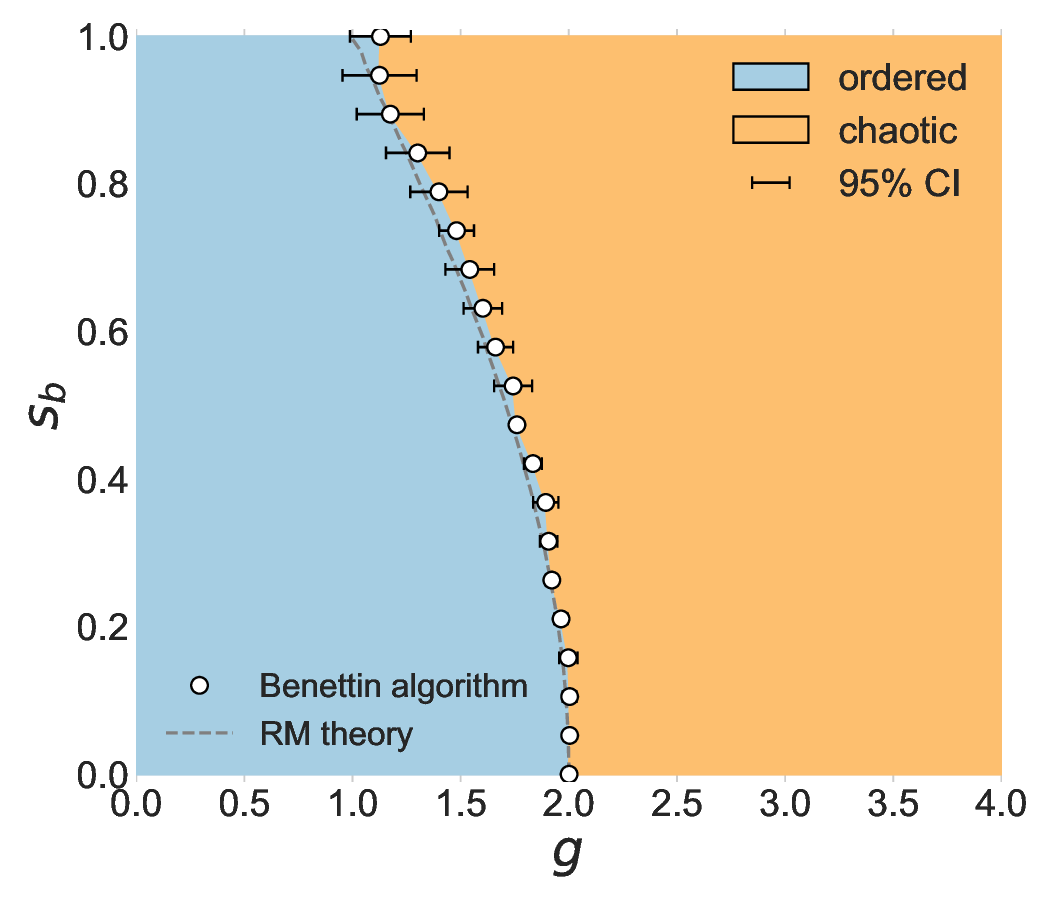}
  \end{subfigure}
 
  \vspace{0.8em}
 
  \begin{subfigure}[t]{\linewidth}
    \centering
    \includegraphics[width=0.85\linewidth]{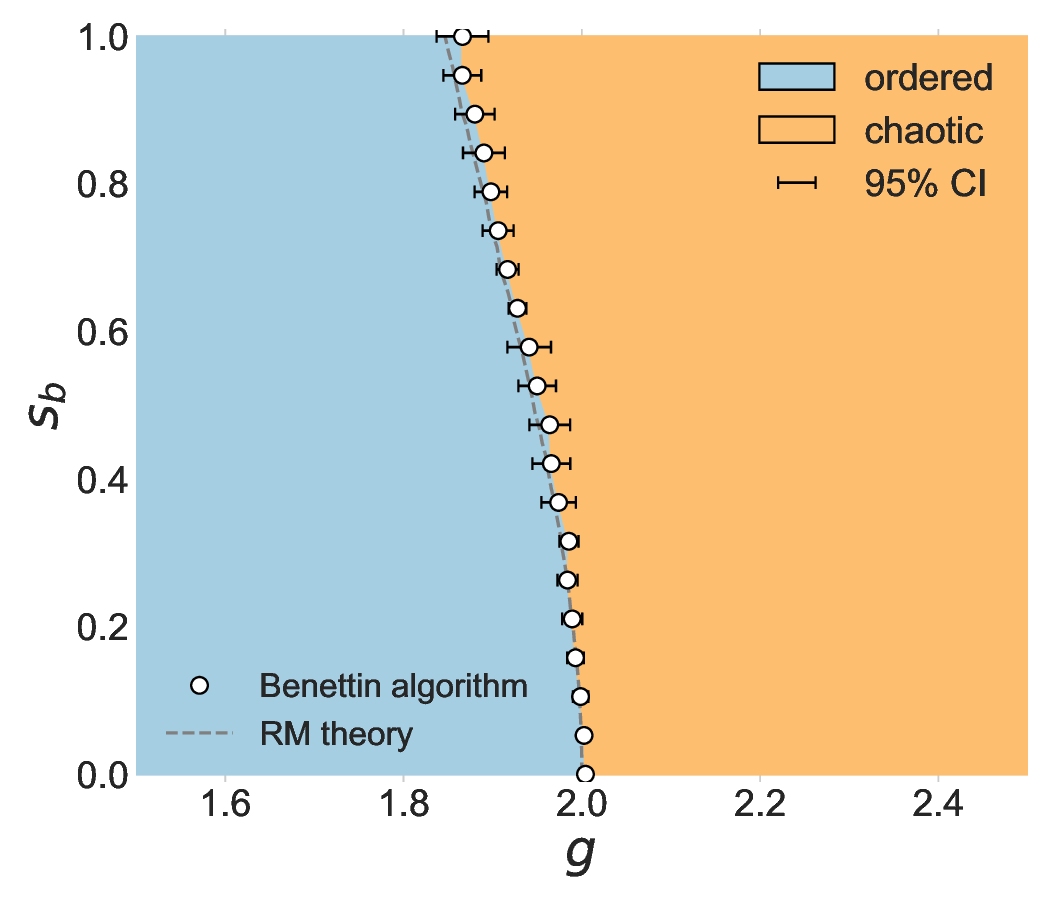}
  \end{subfigure}
 
  \caption{Phase diagram under Gaussian bias initialization for LSTM (top) and GRU (bottom). Dots mark the values of $g$ at which the Benettin estimate of $\lambda_{\max}$ crosses zero for each $s_b$; horizontal error bars denote 95\% confidence intervals across independent replicas of the bias and weight samples. The dashed curve is the analytical prediction $g_c(s_b)$ from~\eqref{eq:g_gauss}.}
  \label{fig:gaussian_phase_transition}
\end{figure}

For the LSTM, \eqref{eq:g_gauss} does not admit a closed form as simple as in the zero-bias case, but it becomes deterministic in the limit $N\to\infty$; its dependence on $s_b$ is plotted as the dashed curve in the top panel of Figure~\ref{fig:gaussian_phase_transition}. For the GRU, the simplified expression~\eqref{eq:gc_gru_gauss} yields the dashed curve in the bottom panel. In both cases we observe very good agreement with the numerically estimated phase transition.

Figure~\ref{fig:gaussian_phase_transition} shows that in both architectures $g_c(s_b)$ is a monotonically decreasing function of $s_b$.
The physical origin of this behavior is most transparent in the GRU. Specializing the Jacobian~\eqref{eq:jacobian} to the GRU via Table~\ref{tab:special_cases} gives
\[
J = I - D_{z^*} + g\, D_{z^*}\, U\, D_{r^*},
\]
where $(D_{z^*})_{jj} = \sigma(b_{z,j})$ and $(D_{r^*})_{jj} = \sigma(b_{r,j})$.
The collapse of~\eqref{eq:g_gauss} reflects the fact that the spectral boundary of $J$ reaches the unit circle at a location determined solely by $g\,U\,D_{r^*}$.
The diagonal matrix $D_{r^*}$ rescales each column of $U$ by $\sigma(b_{r,j})$, giving the renormalized weights a variance
\[
\bigl\langle \bigl(U_{ij}\,\sigma(b_{r,j})\bigr)^2 \bigr\rangle
= \frac{1}{N}\,\langle \sigma(b)^2 \rangle,
\]
where
\begin{equation}
\langle \sigma(b)^2 \rangle
= \int_{-\infty}^{+\infty}
\frac{\sigma(b)^2}{\sqrt{2\pi s_b^2}}\,
\exp\!\left(-\frac{b^2}{2 s_b^2}\right) db.
\label{eq:sigma_sq_avg}
\end{equation}
One can show (see Appendix~\ref{app:monotonicity}) that $\langle \sigma(b)^2 \rangle$ is a strictly increasing function of~$s_b$.
Thus, increasing the bias variance increases the variance of the effective renormalized weights, which in turn lowers the critical gain $g_c(s_b)$ needed to drive the system chaotic. A similar mechanism is at work in the LSTM, but the expressions are more involved because no analogous cancellation between $M$ and $L$ occurs.

\subsection{Chrono initialization}
\label{sec:chrono}

We now turn to the chrono initialization~\cite{Tallec2018}, a popular scheme for LSTMs.
The prescription samples the forget-gate bias as
\[
b_{f,i} \sim \log\!\bigl(\mathcal{U}([1, T_{\max}-1])\bigr)
\]
and ties the input-gate bias to it via $b_{i,i} = -b_{f,i}$, where $T_{\max}$ is a hyperparameter setting the longest timescale the network is expected to capture; the output-gate bias $b_o$ is left unspecified.

Setting $\tau_i := 1 + T_i$ with $T_i \sim \mathcal{U}(1, T_{\max}-1)$, so that $\tau_i \sim \mathcal{U}(2, T_{\max})$, the prescription reads
\[
b_{f,i} = \log(\tau_i-1), \qquad b_{i,i} = -\log(\tau_i-1).
\]
At the autonomous fixed point $\underline{h} = 0$, every gate reduces to $\sigma(b_{\mathrm{g}, i})$, giving
\[
f^*_i = \sigma\bigl(\log(\tau_i-1)\bigr) = \frac{\tau_i-1}{\tau_i}, \quad
i^*_i = \sigma\bigl(-\log(\tau_i-1)\bigr) = \frac{1}{\tau_i}.
\]
In particular $f^*_i + i^*_i = 1$. Using the LSTM entries $A = f+i$, $\alpha = i/(f+i)$ of Table~\ref{tab:special_cases}, we find $A^*_i = 1$ and $\alpha^*_i = 1/\tau_i$, so that
\[
M = I - \operatorname{Diag}(1/\tau_i), \quad
L = \operatorname{Diag}(1/\tau_i), \quad
R = \operatorname{Diag}(\sigma(b_{o,i})).
\]

The key structural consequence is that $L = I - M$ at the fixed point. This is precisely the identity that the GRU satisfies by construction: its single update gate $z_t$ plays simultaneously the role of $\alpha$ and $1-\alpha$ via
\[
\underline{h}_{t+1} = (1-z_t)\,\underline{h}_t + z_t\,\phi(\hat{c}_{t+1}),
\]
so that $L$ and $I - M$ coincide as $\operatorname{Diag}(\sigma(b_z))$. In the LSTM, $L$ and $I-M$ generically involve two independent bias vectors $b_i$ and $b_f$, but the chrono prescription enforces $\sigma(b_i) = 1 - \sigma(b_f)$ entry by entry, thereby reproducing the GRU's structural symmetry at $\underline{h}=0$.

Once $L = I - M$, the ratio $L_{ii}^2/(1-M_{ii})^2 \equiv 1$ pointwise, and the $\tau_i$-dependent factors cancel identically inside the sum in~\eqref{eq:critical_g}, which collapses to
\begin{equation}
g_c = \left(\frac{1}{N}\sum_{i=1}^{N} \sigma(b_{o,i})^2\right)^{-\frac{1}{2}} \xrightarrow[N\to\infty]{} \bigl\langle \sigma(b_o)^2 \bigr\rangle^{-1/2}.
\label{eq:g_crono}
\end{equation}
Crucially, the critical gain does not depend on the chrono timescales: neither the individual $\tau_i$'s, nor their distribution, nor the hyperparameter $T_{\max}$ enter~\eqref{eq:g_crono}---they are absorbed into $L$ and $M$ and cancel against each other. What remains is a closed-form expression in the output-gate bias alone, structurally identical to the GRU result~\eqref{eq:gc_gru_gauss}, with $b_o$ playing the role of the GRU reset-gate bias $b_r$. In particular, the common choice $b_o = 0$ yields $g_c = 2$ regardless of how the timescale range is chosen, exactly as for a zero-reset-bias GRU.
\section{Reservoir computing}
\label{sec:rc}

In this section we consider a reservoir-computing setup using 
the gated RNN as a reservoir. The recurrent weights and biases 
are sampled once at initialization and then kept fixed; only a 
ridge regressor is trained to predict the chaotic 
Mackey--Glass time series~\cite{MackeyGlass1977} using the 
hidden state as feature vector. We find that, consistently with the EOC hypothesis in the reservoir-computing literature, test performance is near-optimal when the gain is set close to the critical value predicted by~\eqref{eq:critical_g}.

The Mackey--Glass system is defined by
\begin{equation}
u(t + 1) = (1 - \gamma)\, u(t) 
+ \beta \,\frac{u(t - \tau)}{1 + u(t - \tau)^n}
\label{eq:mackey_glass}
\end{equation}
with $\beta = 0.2$, $\gamma = 0.1$, $n = 10$, and 
$\tau = 25$, which yield chaotic dynamics. The choice of this 
time series is motivated by its widespread use in the 
literature ~\cite{Jaeger2001, Jaeger2004}.

\begin{figure}[t!]
    \centering
    \includegraphics[width=\linewidth]{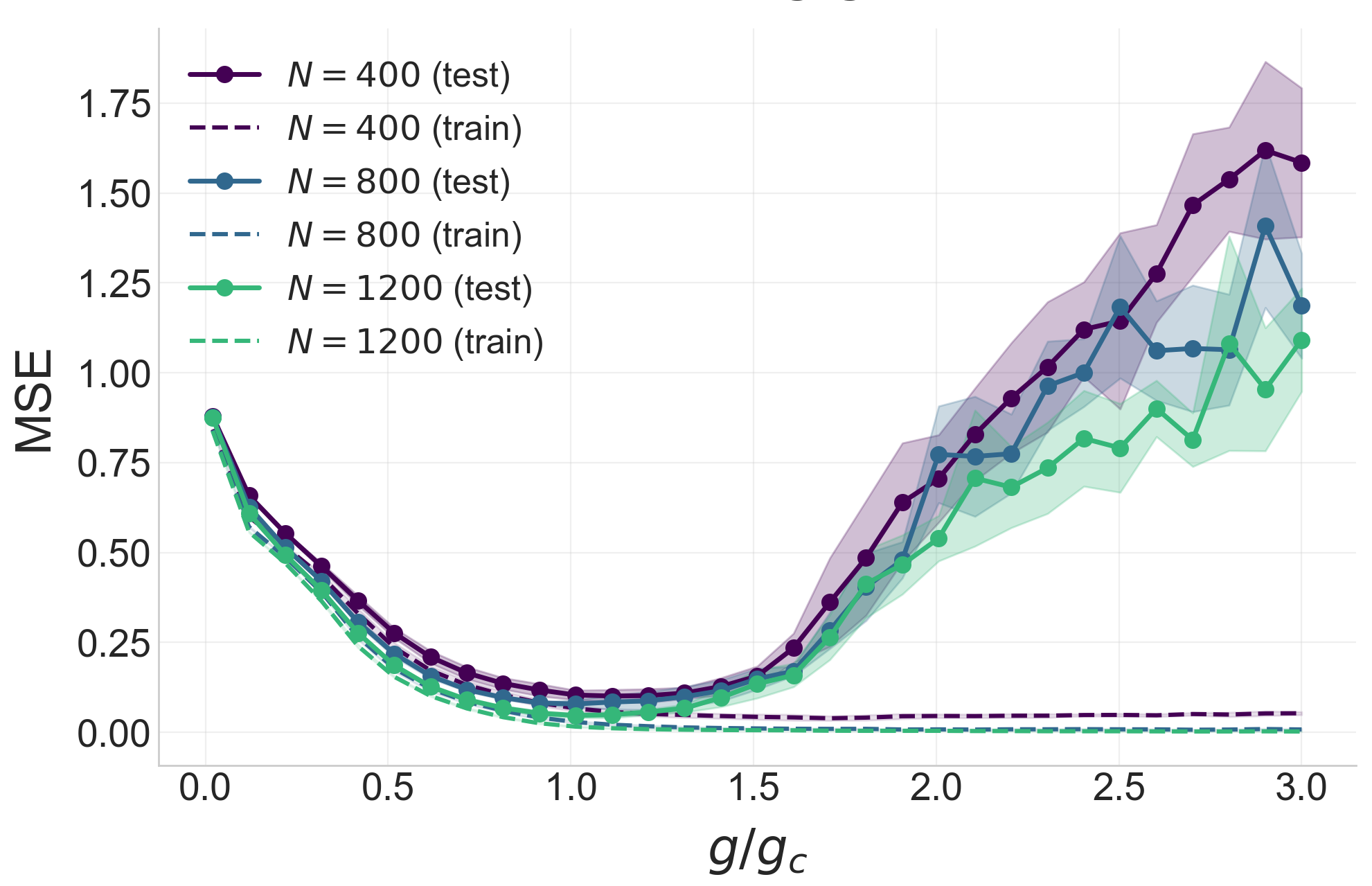}
    \caption{Training (dashed) and test (solid) mean squared error for Mackey--Glass prediction with zero-bias LSTM reservoirs, as a function of the normalized gain $g/g_c$, for different reservoir sizes $N$. The test MSE exhibits a clear minimum close to $g/g_c = 1$ and the training MSE instead decreases monotonically, saturating near zero in the chaotic regime $g/g_c > 1$ (overfitting).}
    \label{fig:mse_RC}
\end{figure}

When the series to predict is chaotic, the reservoir has to both remember a sufficient amount of information from the input history and be sufficiently expressive to transform the input into a rich representation, which is exactly what happens at the EOC.
Thus, it has been hypothesized and empirically observed that initializing the reservoir at the EOC leads to optimal performance~\cite{Bertschinger2004,Boedecker2012}.
In this framework, we can thus verify if there is agreement between the optimal performance and the critical gain predicted by our tool.

We first simulated reservoirs using LSTMs initialized with zero bias and studied the training and test loss as a function of the ratio $g/g_c$ for different reservoir sizes $N$, to both verify the theoretical prediction and check for finite-size effects.
The results reported in Figure~\ref{fig:mse_RC} show a clear minimum of the mean squared error (MSE) at a gain value that is in very good agreement with the critical gain predicted by the instability of the trivial fixed point. The minimum is in fact attained at a gain that is only minimally larger than the theoretical critical value, a small systematic shift that can be attributed to the presence of an external input, which is not accounted for in the theoretical analysis and tends to suppress the chaotic dynamics.
Although we did not investigate this effect systematically, we empirically observed that, as the input amplitude is reduced, the prediction accuracy improves and the optimal performance approaches the gain closest to the theoretical critical value; upon further decreasing the input amplitude, the performance starts to degrade.

We also see that the MSE is lower for larger reservoirs, indicating that for the sizes considered the regression is not overfitting yet and the higher dimensionality of the feature space helps in improving the performance.

Note also that the training loss does not show a minimum at the critical gain, but has instead a monotonically decreasing behavior, approaching zero in the chaotic regime.
This is expected, as in the chaotic regime the reservoir is sufficiently expressive to memorize the training set but fails to generalize due to the sensitivity to initial conditions, i.e.\ the regression is overfitting.

\begin{figure}[t!]
    \centering
    \includegraphics[width=\linewidth]{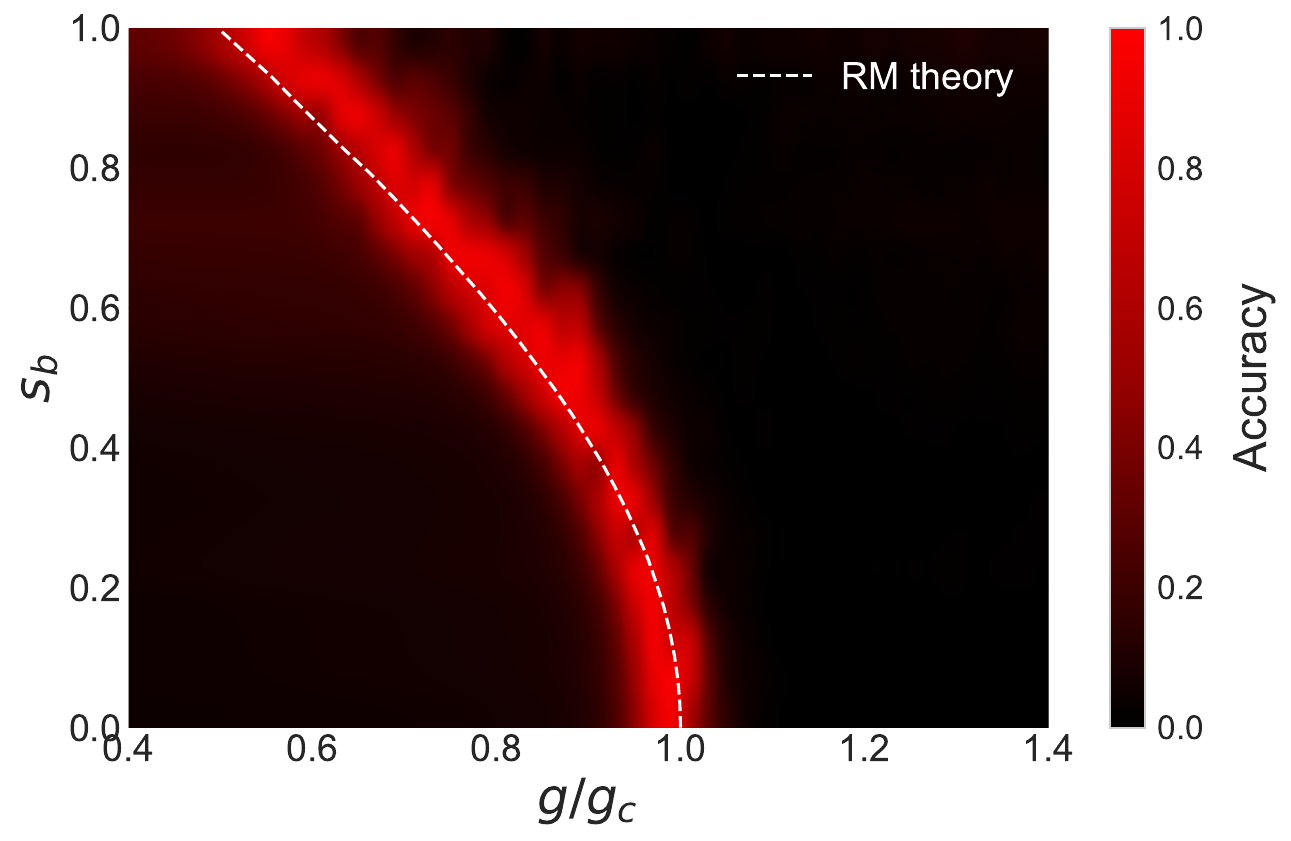}
    \caption{Heatmap of reservoir-computing performance for Mackey--Glass prediction using an LSTM reservoir under Gaussian initialization. Accuracy is measured as $1/\mathrm{Test\ MSE}$ and then normalized row-wise (each row rescaled to lie in $[0,1]$), so that the color indicates the location of the optimum along $g/g_c$ at fixed $s_b$. The white dashed curve shows $g_c(s_b)/g_c$ as predicted by \eqref{eq:g_gauss}.}
    \label{fig:rc_heatmap}
\end{figure}

Lastly, we fixed the reservoir size to $N = 2000$ and extended the analysis to the case of varying bias standard deviation $s_b$, where all biases were drawn from a Gaussian distribution.  
The results, reported in Figure~\ref{fig:rc_heatmap}, show again a clear peak of performance along the theoretical critical line predicted by our analysis, confirming that initializing the reservoir at the EOC leads to optimal performance even in the biased case.

\section{Discussion and Limitations}
\label{sec:discussion}
In this work we have treated gated recurrent neural networks as 
high-dimensional random dynamical systems undergoing a phase 
transition controlled by the weight variance~$g^2$, in the spirit 
of the seminal analysis of Sompolinsky \emph{et al.}~\cite{Sompolinsky1988}. 
Guided by the observation that, in that model, the onset of chaos 
coincides with the instability of the zero fixed point, we have 
shown numerically that the same correspondence holds for a broader 
class of gated architectures including LSTMs and GRUs, across 
different bias initializations. 
Leveraging this correspondence and the spectral theory 
of~\cite{Fumarola2015}, we have derived a closed-form 
criterion~\eqref{eq:critical_g} for the critical gain~$g_c(s_b)$ as a 
function of the bias, and validated it against Lyapunov-exponent 
estimates. In a reservoir-computing setting on the chaotic 
Mackey--Glass time series, test performance peaks at the predicted $g_c$, providing a concrete illustration of how the criterion can be used in practice.
 
Our approach is built on the observation that the fixed-point instability provides a reliable proxy for the EOC, which we verify numerically across all the architectures and initialization schemes considered. While we do not prove this correspondence in full generality and refer the reader to~\cite{Can2020, Krishnamurthy2022} for a more theoretical treatment, it is sufficient to derive a closed-form criterion that can be readily used by practitioners.
 
Two simplifications underlie our analysis. 
The first concerns the structure of the fixed point: removing the 
bias from the candidate update enforces~$\underline{h}=0$ as a fixed point, 
making the linearization possible. 
The second is a parameterization choice: sampling all gate weight 
matrices with the same gain ensures that only the candidate 
matrix enters the linearized Jacobian. 
Relaxing the first assumption---introducing a candidate 
bias---would break the symmetry around the origin, potentially 
leading to neuron-dependent fixed points and a less tractable mathematical framework.
Relaxing the second---allowing independent gains per 
gate---would open up additional degrees of freedom that could 
give rise to qualitatively different dynamical regimes and a richer phase diagram
(see~\cite{Krishnamurthy2022} for a deeper 
analytical investigation). 
On the representational side, the zero-bias symmetry would 
constrain vanilla RNNs in the reservoir computing setting to odd input--output maps; however, in 
gated architectures the multiplicative gating structure breaks 
this oddness, and to the authors' knowledge, it remains an open question whether the 
constraint reduces in practice the expressivity of the gated network.
 
Finally, recent work by Cowsik \emph{et al.}~\cite{Cowsik2025} has shown that an analogous 
order-to-chaos transition, governed by initialization 
hyperparameters, predicts trainability also in deep transformers. 
This suggests that criteria such as~\eqref{eq:critical_g} may be 
relevant beyond reservoir computing, serving as reference points 
for the initialization of networks whose weights are subsequently 
trained, a direction we plan to explore in future work.
\begin{acknowledgments}
We thank Jean-Philippe Bouchaud, Eric Vanden-Eijnden and in particular Giulio Biroli  
for their precious comments and suggestions. 
\end{acknowledgments}
\bibliographystyle{apsrev4-2}
\bibliography{refs}

\appendix
\onecolumngrid
\section{Removing the regularizer}
\label{app:reg}

The boundary condition~\eqref{eq:fumarola_condition} is taken from~\cite{Fumarola2015}
in its full form, with the limits in the order $\lim_{r\to 0^+}\lim_{N\to\infty}$.
This order matters in general: for arbitrary deterministic sequences of $M, L, R$,
the unregularized empirical sum may diverge as $N\to\infty$ if a sufficient fraction
of the singular values $\mathrm{sv}_i(z)$ vanishes fast enough. The role of the regularizer $r>0$ is 
to prevent the $N$-limit to diverge and extend the formula to these pathological cases.

In this appendix we argue that, for the bias initializations of
Section~\ref{sec:experiments}, the regularized condition~\eqref{eq:fumarola_condition}
and the unregularized one~\eqref{eq:critical_g} yield the same prediction for the
critical gain. As stated in Section~\ref{sec:theory},
we do not attempt to verify all the technical conditions of~\cite{Fumarola2015} but
the following argument settles only the change of limits issue.

As noted in the main body, the relevant locus is the point on the unit circle that the boundary of the
spectral support touches first. Since $M_{ii}\in[0,1)$ in all cases considered,
\begin{equation}
    |z - M_{ii}|^2 - (1 - M_{ii})^2 \;=\; 2\,M_{ii}\bigl(1-\operatorname{Re}(z)\bigr) \;\geq\; 0
    \qquad \text{for } |z|=1,
\end{equation}
so $\mathrm{sv}_i(z) \geq \mathrm{sv}_i(1)$ pointwise on the unit circle and it
is enough to control the expectation $\langle 1/\mathrm{sv}(1)^2 \rangle$.
For each $r>0$ the variables $1/(\mathrm{sv}_i(1)^2+r^2)$ are iid and bounded by $1/r^2$, so the strong law of large numbers gives, almost surely,
\begin{equation}
    \frac{1}{N}\sum_{i=1}^{N}\frac{1}{\mathrm{sv}_i(1)^2+r^2}
    \;\xrightarrow[N\to\infty]{}\;
    \left\langle\frac{1}{\mathrm{sv}(1)^2+r^2}\right\rangle,
\end{equation}
and monotone convergence as $r\to 0^+$ raises the right-hand side to $\langle 1/\mathrm{sv}(1)^2\rangle$ (recall that $1/\mathrm{sv}_i(1)^2 = L_{ii}^2 R_{ii}^2 / (1-M_{ii})^2$). The two orders of limits therefore agree whenever this last expectation is finite, which we now check in the three cases of Section~\ref{sec:experiments}.
\subsection*{Zero bias}
All entries are deterministic and equal to $1/2$, so $1/\mathrm{sv}_i(1)^2 = 1/4$
and $\langle 1/\mathrm{sv}(1)^2 \rangle = 1/4$.

\subsection*{Gaussian bias}
For the GRU, $L = I - M$ holds pointwise (cf.~Table~\ref{tab:MLR}), so the ratio
$L_{ii}^2/(1-M_{ii})^2$ is identically $1$ and $1/\mathrm{sv}_i(1)^2 = R_{ii}^2 \leq 1$,
hence $\langle 1/\mathrm{sv}(1)^2 \rangle \leq 1$.
For the LSTM the biases $b_f, b_i, b_o$ are independent and no such cancellation
occurs. The factors $L_{ii}^2$ and $R_{ii}^2$ are bounded by one, and the only
nontrivial term is the expectation of $(1-\sigma(b_f))^{-2}$. Using
$1-\sigma(x) = 1/(1+e^x)$ and the moment generating function of a Gaussian,
\begin{equation}
    \bigl\langle (1-\sigma(b_f))^{-2}\bigr\rangle
    \;=\; \bigl\langle (1+e^{b_f})^2\bigr\rangle
    \;=\; 1 + 2\,e^{s_b^2/2} + e^{2 s_b^2},
\end{equation}
which is finite for any finite $s_b$. By independence,
$\langle 1/\mathrm{sv}(1)^2 \rangle$ is bounded by the same quantity.

\subsection*{Chrono initialization}
The relations $L = \operatorname{Diag}(1/\tau_i)$ and $M = I - \operatorname{Diag}(1/\tau_i)$
of Section~\ref{sec:chrono} again give $L = I - M$ pointwise, so as in the GRU case
$1/\mathrm{sv}_i(1)^2 = R_{ii}^2 = \sigma(b_{o,i})^2 \leq 1$, hence
$\langle 1/\mathrm{sv}(1)^2 \rangle \leq 1$, independently of the distribution
of the timescales $\tau_i$.
\section{Monotonicity of \texorpdfstring{$F(s_b)$}{F(sb)}}
\label{app:monotonicity}
\noindent We show that the function
\begin{equation}
  F(s_b) \;:=\; \bigl\langle \sigma(b)^{2} \bigr\rangle
  = \bigl\langle \sigma(s_b\, z)^{2} \bigr\rangle_{z},
  \qquad z \sim \mathcal{N}(0,1),
\end{equation}
is strictly increasing for every $s_b > 0$.
Differentiating under the integral sign yields
\begin{equation}
  F'(s_b)
  = 2\,\bigl\langle \varphi(s_b\, z)\, z \bigr\rangle_{z},
  \qquad
  \varphi(u) := \sigma(u)^{2}\,\bigl[1 - \sigma(u)\bigr].
\end{equation}
Splitting the expectation into contributions from $z > 0$ and $z < 0$
and exploiting the even symmetry of the standard Gaussian density
$p(z) = (2\pi)^{-1/2}\, e^{-z^{2}/2}$, we obtain
\begin{equation}
  F'(s_b)
  = 2 \int_{0}^{\infty}
    \bigl[\varphi(s_b\, z) - \varphi(-s_b\, z)\bigr]\, z\, p(z)\, dz.
\end{equation}
Using the identity $\sigma(-u) = 1 - \sigma(u)$, a direct computation gives
\begin{equation}
  \varphi(u) - \varphi(-u)
  = \sigma(u)\,\bigl[1 - \sigma(u)\bigr]\,\bigl[2\,\sigma(u) - 1\bigr].
\end{equation}
For $u > 0$ one has $\sigma(u) \in \bigl(\tfrac{1}{2},\, 1\bigr)$,
so each of the three factors is strictly positive.
The integrand is therefore strictly positive for all $z > 0$,
which implies $F'(s_b) > 0$ for every $s_b > 0$. 

\twocolumngrid

\end{document}